\documentclass[a4paper]{esannV2}
\usepackage[dvips]{graphicx}
\usepackage[latin1]{inputenc}
\usepackage{amssymb,amsmath,array}

\usepackage{color}
\usepackage{epstopdf}
\usepackage[export]{adjustbox}
\usepackage{hyperref}

\begin{document}
\title{Investigating the Gestalt Principle of Closure in Deep Convolutional Neural Networks}

\author{Yuyan Zhang$^1$, Derya Soydaner$^{2,3}$, Fatemeh Behrad$^{2,3}$, Lisa Ko{\ss}mann$^2$, \\ and Johan Wagemans$^{2,3}$
%
\thanks{This work was supported by the European Research Council (ERC) for the `Gestalts Relate Aesthetic Preferences to Perceptual Analysis'
(GRAPPA) Project under Grant 101053925.}
%
\vspace{.3cm}\\
%
1- KU Leuven, Department of Computer Science, 
Leuven, Belgium
%
\vspace{.1cm}\\
2- KU Leuven, Department of Brain and Cognition, 
Leuven, Belgium
\vspace{.1cm}\\
3- Leuven.AI, KU Leuven Institute for AI, Leuven, Belgium \\
}

\maketitle

\begin{abstract} Deep neural networks perform well in object recognition, but do they perceive objects like humans? This study investigates the Gestalt principle of closure in convolutional neural networks. We propose a protocol to identify closure and conduct experiments using simple visual stimuli with progressively removed edge sections. We evaluate well-known networks on their ability to classify incomplete polygons. Our findings reveal a performance degradation as the edge removal percentage increases, indicating that current models heavily rely on complete edge information for accurate classification. The data used in our study is available on \href{https://github.com/zhangyy708/closure-in-CNNs}{Github}.

\end{abstract}

\section{Introduction}
Neural networks were designed inspired by the working mechanism of the human brain and have since achieved remarkable success across various fields. While psychology still aims to better understand the human brain, computer science strives to enhance understanding of neural networks. The primary goal in neural network research is to develop models capable of performing similar tasks as the human brain, rather than to recreate it. Interestingly, neural networks tend to exhibit more human-like behavior than expected, despite not being explicitly designed for this purpose. In particular, recent findings suggest that CNNs may exhibit certain aspects of Gestalt laws of perceptual organization \cite{wertheimer1923}, which explain how the human brain interprets complex visual stimuli, albeit possibly subject to certain thresholds and limitations.

The narrow scope of previously explored neural network architectures, coupled with limited datasets and insufficient experiments, necessitates a more comprehensive investigation. We focus on the Gestalt principle of closure, which states that the human brain naturally fills in gaps to perceive figures as complete wholes, when parts are occluded or fragmented. We present a dataset specifically designed to examine closure from various psychology-based perspectives and conduct experiments across a broad range of CNNs to investigate their alignment with this principle. Our work provides a comprehensive analysis of CNNs regarding closure, identifying limits and thresholds that define their applicability in performing closure over gradually manipulated stimulus classes. 

\section{Preliminaries}

\subsection{Gestalt laws of perceptual organization}
Human visual perception has multiple, hierarchical processing stages. At the low-level processing stages, the visual system encodes features like luminance and color, at the mid-level stages elements are grouped and organized, and at the high-level stages tasks of identifying and understanding are performed. One significant contribution to research on visual perception was the formulation of the principles of perceptual grouping, often also referred to as ``Gestalt Laws". Grouping describes the phenomenon that observers perceive certain elements as belonging to each other compared to others, which appear distinct. The classic laws of grouping are: proximity, similarity, common fate, good continuation, symmetry, parallelism and closure \cite{wagemans2012}.

\subsection{Law of closure}
The law of closure refers to a shared human preference for closed shapes over disjoined individual elements. This leads the visual system to complete contours and to create closed shapes. Depending on the elements and mechanisms at play, this can be referred to as contour integration or contour completion. Contour integration occurs when distinct elements such as dots or lines are integrated into the contour of a shape \cite{wagemans2012}. Contour completion refers to the integration of smooth contours \cite{wagemans2012}. If a shape or scene is completed behind an occluder, this is referred to as amodal completion. If the perception of an illusory contour or surface is triggered by stimulus characteristics, this can be referred to as modal completion \cite{wagemans2006}. The distinction between the two is made by the presence or absence of certain visual qualities in the completion \cite{wagemans2006}.

\section{Related Work}

Amanatiadis \emph{et al.} \cite{amanatiadis2018} trained AlexNet \cite{krizhevsky2017imagenet} and Inception V1 (GoogLeNet) \cite{szegedy2015going} on the MNIST \cite{lecun1998} and ImageNet \cite{deng2009} datasets to explore six core Gestalt laws. They measured closure as a function of occlusion percentage and found that the closure principle is effective in the CNNs up to approximately 30\% occlusion, beyond which the models' performance decreases. Another study examined a wider range of models including ResNet-152 \cite{7780459} and DenseNet-201 \cite{huang2017densely}, and found mixed evidence of perceptual grouping \cite{biscione2023}. Ehrensperger \emph{et al.} \cite{ehrensperger2019} analyzed how AlexNet and Inception V1 classify Kanizsa triangles (a famous example of modal completion) and modified triangles with sections of the edges removed, reporting closure. Kim \emph{et al.} \cite{kim2021} showed that both Inception and a simple CNN, trained to classify natural images, exhibit closure on synthetic displays of edge fragments. They tested these CNNs on incomplete triangles and demonstrated that they were more likely to be recognized as similar to complete triangles rather than disordered fragments. Baker \emph{et al.} \cite{baker2018deep} fine-tuned AlexNet for a shape discrimination task, training it to classify wireframes and Kanizsa squares as `fat' and `thin'. In contrast to earlier work, they concluded that neural networks do not perceive illusory contours. With limited stimuli and metrics employed on only a few CNNs, making a definitive statement about the presence of closure is not possible. Our method trains on complete polygons and tests on incomplete ones. Our approach, along with our examination of a broader range of CNN models, provides a more comprehensive assessment of closure performance in CNNs.

\section{Convolutional neural networks}
CNNs have greatly advanced computer vision by effectively learning spatial features from images through a series of convolutional layers. Notable CNN architectures have emerged, each characterized by unique design choices and strengths. Early CNN models like AlexNet \cite{krizhevsky2017imagenet} and VGG16 \cite{simonyan2014very} achieved impressive accuracy but suffered from high computational complexity due to their reliance on numerous small convolutional filters stacked together. Recent advancements, exemplified by architectures such as ResNet, SqueezeNet \cite{iandola2016squeezenet}, and DenseNet, have addressed this challenge by introducing innovative techniques such as residual connections, and efficient filter design.

Moreover, Inception network \cite{szegedy2015going} and ShuffleNet \cite{zhang2018shufflenet} improve accuracy by incorporating filters of varying sizes within a single layer and leveraging channel shuffle operations, respectively. EfficientNet \cite{tan2019efficientnet} employs a compound scaling method to optimize network depth, width, and resolution according to specific resource constraints. MobileNet \cite{howard2017mobilenets}, with its lightweight design and depth-wise separable convolutions, is particularly suitable for deployment on mobile devices.

In our experiment, we freeze all layers of the mentioned CNNs. The last layer is replaced with a fully-connected layer with 10 nodes corresponding to 10 different polygons in the dataset. The networks are fine-tuned for 25 epochs with a batch size of 32. We use the Adam optimizer \cite{kingma2014} with a learning rate of .001 and cross-entropy loss function.

\section{Experimental results}

\subsection{Dataset}
\begin{figure}[b!]
\includegraphics[trim=0 50 0 80pt, clip, scale=0.50, center]{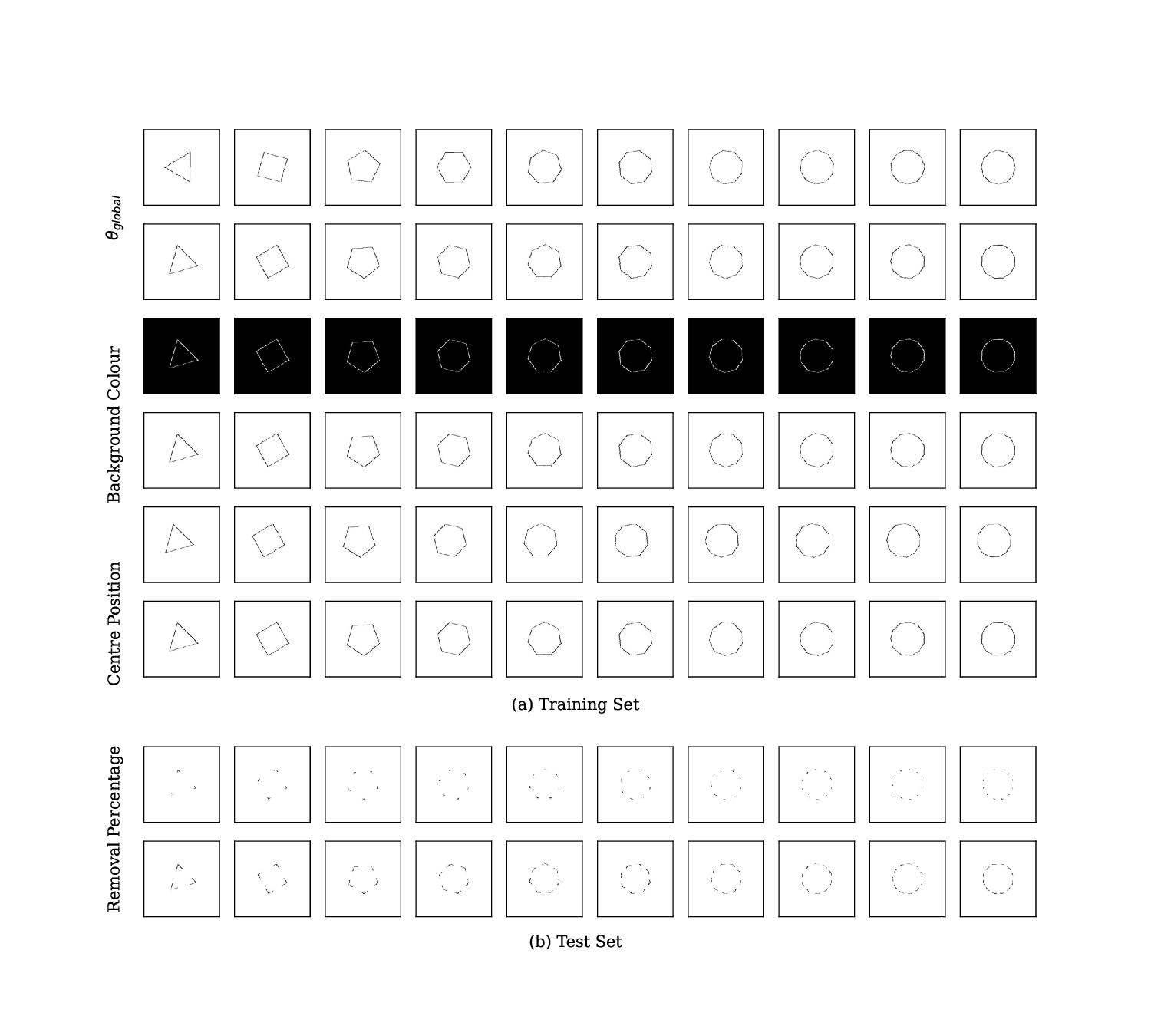}
\caption{Image examples used in the \emph{(a)} training set \emph{(b)} test set. Each column indicates a different type of polygon (i.e., triangle, square, pentagon, hexagon, and so on). In (a), the top two rows show polygons with different $\theta_{global}$, the middle two rows feature different background colours, and the bottom two rows illustrate various positions. In (b), each row shows polygons with varying percentages of removal from each side. The settings for $\theta_{global}$, background colours, and centre positions of the polygons in the test set are the same as those used in the training set, although these variables are not depicted in (b). }
\label{fig:data4}
\end{figure}

For the experiment, we generated a novel dataset with a training set (Fig.~\ref{fig:data4} (a)) consisting of 320 complete polygons, which vary in the number of sides (3 to 12), $\theta_{global}$ (defined as the degree to which the polygon rotates around its centre in the plane; 0\textdegree, 15\textdegree, 30\textdegree, 45\textdegree, 60\textdegree, 75\textdegree, 90\textdegree, and 105\textdegree), background colours (white or black), and centre positions (the centre of the canvas or 16 pixels to the left and 16 pixels to the bottom of the canvas' centre). Assuming the polygon has $N$ sides, then its $\theta_{global}$ could have the value of  $180(N-2) * i / N$ \textdegree ($i = 0, 1, 2, 3, 4, 5, 6, 7$). 

The test set (Fig.~\ref{fig:data4} (b)) for the classification includes polygons which differ from each other in the removal percentages. This refers to the space between the line fragments that potentially form the contour of the shape. The removal percentage of a polygon could be 0 (a complete polygon), 10, 20, 30, 40, 50, 60, 70, 80, or 90. A removal percentage means that on each side of a polygon section of the contour has been removed (or become invisible) and the invisible part is always in the middle of each side. This setting will result in 10 test sets (based on the removal percentage) each including 320 images.

\subsection{Closure Measurement}
We employ the percentage of the correctly classified instances in the test set ($p$) as the indicator of the closure effect. A stable classification performance through the different removal percentage stages above 10\% (randomly selecting a class for a given instance) would indicate evidence for the existence of the closure effect in the model. 

\subsection{Results and discussion}
\begin{figure}[htbp]
\includegraphics[scale=0.45, center]{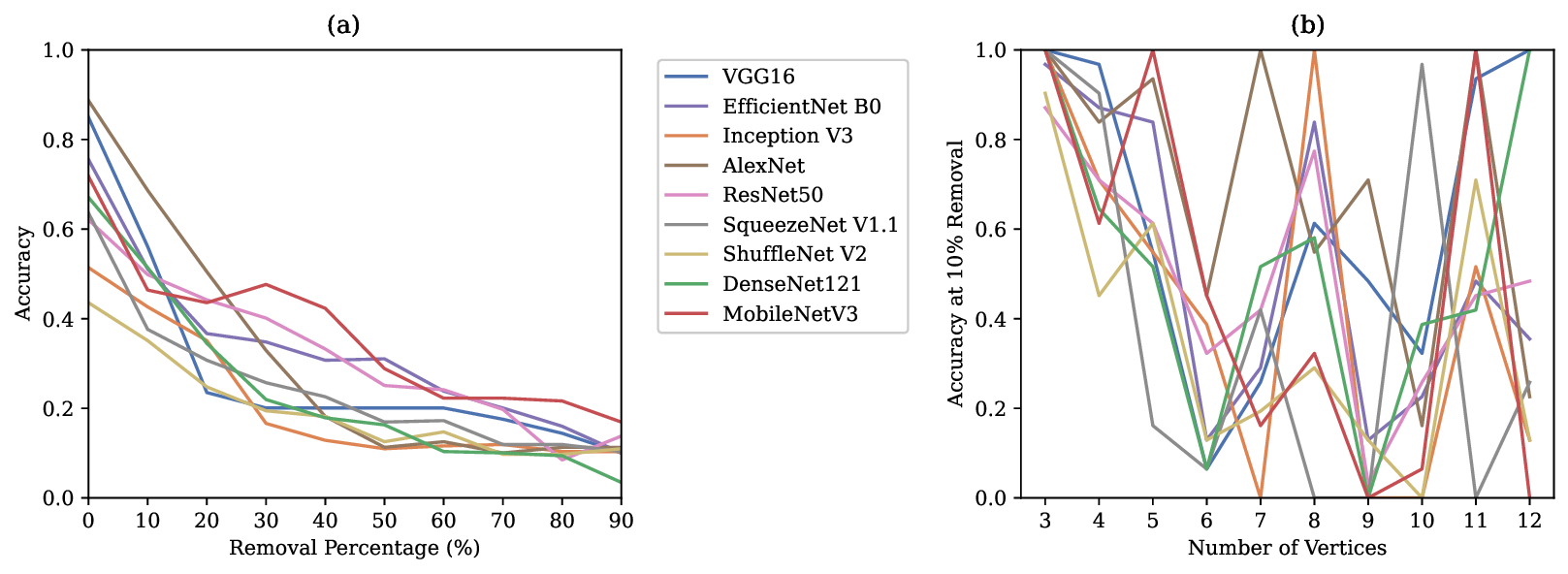}
\caption{The percentage of correctly classified examples in the test set versus \emph{(a)} the removal percentage for each model in each side of a polygon and \emph{(b)} the number of vertices at 10\% removal percentage. }
\label{fig:rst4}
\end{figure}

The performance of models on the training set (when the removal percentage is 0) is depicted in Fig. \ref{fig:rst4}. VGG16 and SqueezeNet V1.1 perform relatively well with an accuracy of around 90\%. AlexNet and ResNet50 also have a good performance and achieve an accuracy of higher than 70\%. Inception V3, ShuffleNet V2 and DenseNet121 obtain an accuracy of more than 60\%. However, the accuracy of EfficientNet B0 and MobileNetV3 are between 40\% and 50\%, respectively, due to their lower complexities. It suggests that they cannot do the task reliably.

A closer inspection of model performance reveals that accuracy decreases by removal percentage as a general trend (Fig.~\ref{fig:rst4}(a)), but the performance of individual CNNs over vertices is inconsistent (Fig.~\ref{fig:rst4}(b)). The accuracy of classification remains much higher than chance level (10\%) when the removal percentage is not larger than 20\% to 30\%, although it decreases quickly when the percentage of removal is larger. This aligns with the findings of Amanatiadis \emph{et al.} \cite{amanatiadis2018}. However, through the small gradual manipulations in our experiment, we observe a drastic decrease in accuracy already at the first stimulus manipulation. This is not in accordance with contour completion elicited by the closure effect.

\section{Conclusion}
We demonstrate that the classification performance of CNNs deteriorates with partial removal of the contours in incomplete polygons, by using 10 different polygons with gradually manipulated contours.
Most models perform the classification task above the chance level for removal percentages under 30\%, which could be seen as an indication of performing contour completion. However, classification performance already drastically deteriorates for all networks at only 10\% of contour removal, and is inconsistent across number of vertices, which might instead be symptomatic of CNNs high dependence on local features. In conclusion, CNNs do not display the closure effect with current measurements and datasets.

%





\begin{footnotesize}

\bibliographystyle{unsrt}
\bibliography{references}

\begin{thebibliography}{10}

\bibitem{wertheimer1923}
M.~Wertheimer.
\newblock Laws of organization in perceptual forms.
\newblock {\em Psychologische Forschung}, 4:301--350, 1923.

\bibitem{wagemans2012}
J.~Wagemans et~al.
\newblock A century of gestalt psychology in visual perception: I. perceptual grouping and figure--ground organization.
\newblock {\em Psychological Bulletin}, 138(6):1172, 2012.

\bibitem{wagemans2006}
J.~Wagemans, R.~Van Lier, and B.J. Scholl.
\newblock Introduction to {Michotte's} heritage in perception and cognition research.
\newblock {\em Acta Psychologica}, 123(1-2):1--19, 2006.

\bibitem{amanatiadis2018}
A.~Amanatiadis, V.G. Kaburlasos, and E.B. Kosmatopoulos.
\newblock Understanding deep convolutional networks through gestalt theory.
\newblock In {\em IEEE International Conference on Imaging Systems and Techniques (IST)}, 2018.

\bibitem{krizhevsky2017imagenet}
A.~Krizhevsky, I.~Sutskever, and G.E. Hinton.
\newblock Imagenet classification with deep convolutional neural networks.
\newblock {\em Communications of the ACM}, 60(6):84--90, 2017.

\bibitem{szegedy2015going}
C.~Szegedy et~al.
\newblock Going deeper with convolutions.
\newblock In {\em Proceedings of the IEEE Conference on Computer Vision and Pattern Recognition (CVPR)}, pages 1--9, 2015.

\bibitem{lecun1998}
Y.~LeCun et~al.
\newblock Gradient-based learning applied to document recognition.
\newblock {\em IEEE}, 86(11):2278--2324, 1998.

\bibitem{deng2009}
J.~Deng et~al.
\newblock Imagenet: A large-scale hierarchical image database.
\newblock In {\em IEEE Conference on Computer Vision and Pattern Recognition}, pages 248--255, 2009.

\bibitem{7780459}
K.~He et~al.
\newblock Deep residual learning for image recognition.
\newblock In {\em IEEE Conference on Computer Vision and Pattern Recognition (CVPR)}, pages 770--778, 2016.

\bibitem{huang2017densely}
G.~Huang et~al.
\newblock Densely connected convolutional networks.
\newblock In {\em Proceedings of the IEEE Conference on Computer Vision and Pattern Recognition (CVPR)}, pages 4700--4708, 2017.

\bibitem{biscione2023}
V.~Biscione and J.S. Bowers.
\newblock Mixed evidence for gestalt grouping in deep neural networks.
\newblock {\em Computational Brain \& Behavior}, 6(3):438--456, 2023.

\bibitem{ehrensperger2019}
G.~Ehrensperger, S.~Stabinger, and A.R. S{\'a}nchez.
\newblock Evaluating cnns on the gestalt principle of closure.
\newblock In {\em Artificial Neural Networks and Machine Learning - ICANN 2019: Theoretical Neural Computation}, pages 296--301, 2019.

\bibitem{kim2021}
B.~Kim et~al.
\newblock Neural networks trained on natural scenes exhibit gestalt closure.
\newblock {\em Computational Brain \& Behavior}, 4(3):251--263, 2021.

\bibitem{baker2018deep}
N.~Baker et~al.
\newblock Deep convolutional networks do not perceive illusory contours.
\newblock In {\em Proceedings of the 40th Annual Conference of the Cognitive Science Society}, Madison, WI, 2018. Cognitive Science Society.

\bibitem{simonyan2014very}
K.~Simonyan and A.~Zisserman.
\newblock Very deep convolutional networks for large-scale image recognition.
\newblock {\em arXiv preprint arXiv:1409.1556}, 2014.

\bibitem{iandola2016squeezenet}
F.N. Iandola et~al.
\newblock Squeezenet: Alexnet-level accuracy with 50x fewer parameters and $<$ 0.5 mb model size.
\newblock {\em arXiv preprint arXiv:1602.07360}, 2016.

\bibitem{zhang2018shufflenet}
X.~Zhang et~al.
\newblock Shufflenet: An extremely efficient convolutional neural network for mobile devices.
\newblock In {\em Proceedings of the IEEE Conference on Computer Vision and Pattern Recognition}, pages 6848--6856, 2018.

\bibitem{tan2019efficientnet}
M.~Tan and Q.~Le.
\newblock Efficientnet: Rethinking model scaling for convolutional neural networks.
\newblock In {\em International Conference on Machine Learning}, pages 6105--6114. PMLR, 2019.

\bibitem{howard2017mobilenets}
A.G. Howard et~al.
\newblock Mobilenets: Efficient convolutional neural networks for mobile vision applications.
\newblock {\em arXiv preprint arXiv:1704.04861}, 2017.

\bibitem{kingma2014}
D.P. Kingma and J.~Ba.
\newblock Adam: A method for stochastic optimization.
\newblock {\em arXiv preprint arXiv:1412.6980}, 2014.

\end{thebibliography}

\end{footnotesize}


\end{document}